\documentclass[12pt,english,aps,prl,twocolumn,showpacs,superscriptaddress,footinbib,reprint,noshowpacs]{revtex4-1}
\usepackage{graphicx}
\usepackage{amsmath}
\usepackage{comment}
\usepackage{amssymb}

\usepackage{subfigure}

\usepackage{color}
\usepackage{soul}

\begin{document}


\title{Energy-entropy competition and the effectiveness of stochastic gradient descent in machine learning}

\author{Yao Zhang}
\affiliation{Cavendish Laboratory, University of Cambridge, Cambridge CB3 0HE, United Kingdom}

\author{Andrew M. Saxe}
\affiliation{Center for Brain Science, Harvard University, Cambridge, MA 02138, United States of America}

\author{Madhu S. Advani}
\affiliation{Center for Brain Science, Harvard University, Cambridge, MA 02138, United States of America}

\author{Alpha A. Lee}
\email{aal44@cam.ac.uk}
\affiliation{Cavendish Laboratory, University of Cambridge, Cambridge CB3 0HE, United Kingdom}

\begin{abstract}
Finding parameters that minimise a loss function is at the core of many machine learning methods. The Stochastic Gradient Descent algorithm is widely used and delivers state of the art results for many problems. Nonetheless, Stochastic Gradient Descent typically cannot find the global minimum, thus its empirical effectiveness is hitherto mysterious. We derive a correspondence between parameter inference and free energy minimisation in statistical physics. The degree of undersampling plays the role of temperature. Analogous to the energy-entropy competition in statistical physics, wide but shallow minima can be optimal if the system is undersampled, as is typical in many applications. Moreover, we show that the stochasticity in the algorithm has a non-trivial correlation structure which systematically biases it towards wide minima. We illustrate our argument with two prototypical models: image classification using deep learning, and a linear neural network where we can analytically reveal the relationship between entropy and out-of-sample error.  
\end{abstract}

\makeatother
\maketitle

\section{Introduction}

Methods in machine learning typically involves finding model parameters that minimize a measure of disagreement between model predictions and training data, known as the loss function. Suppose we performed $P$ measurements of independent variables $\mathbf{x}_i \in \mathbb{R}^{N_i}$ and dependent variable $y_i$. Given a parametric model $f(\mathbf{x}_i,\boldsymbol{\theta})$ where $\boldsymbol{\theta} \in \mathbb{R}^N$ are the parameters, we can define a loss function $L(\{\mathbf{x}_i\}_{i=1}^{P},\boldsymbol{\theta})  = R(\boldsymbol{\theta}) + \frac{1}{P}\sum_i l_i(y_i,f(\mathbf{x}_i,\boldsymbol{\theta})) $, where $R$ is a regularising function. The Stochastic Gradient Descent (SGD) algorithm minimises the loss function by going down the direction of the steepest gradient estimated using random subsets of the data \cite{robbins1951stochastic,bottou2010large,bottou2016optimization}:
\begin{equation}
\boldsymbol{\theta}_{t+1} = \boldsymbol{\theta}_{t} - \eta_t \nabla_{\boldsymbol{\theta} }\left( \frac{1}{|B_t|} \sum_{i \in B_t} l_i(\mathbf{x}_i,\boldsymbol{\theta})  + R(\boldsymbol{\theta}) \right),
\label{SGD}
\end{equation}
where we have abbreviated our notation for the loss function by hiding the dependent variables $y_i$, and defined $\eta_t$ as the learning rate, and $B_t$ as a randomly selected batch of training data. Models routinely used in machine learning can contain $N = O(10^{7})$ parameters and are non-convex, thus finding the global minimum is computationally intractable. Nonetheless, it is empirically known that SGD yields state of the art results in machine learning, in the sense that the parameters SGD finds can give accurate predictions outside the training set (so-called generalisation/testing accuracy) \cite{krizhevsky2012imagenet,russakovsky2015imagenet, lecun2015deep,szegedy2016inception,silver2016mastering}. It is highly implausible that SGD finds the global minimum, and it is mysterious why minima that SGD converges to tend to be generalisable.


There is a growing literature that sheds light on this mystery by focusing on the topology of the loss function. For Gaussian random functions, critical points with a loss function much larger than the global minimum are exponentially likely to be saddle points in the high dimensional limit \cite{fyodorov2007replica,bray2007statistics}. Therefore the value of the loss function at most local minima is similar to the global minimum, and some argue that this is why most local minima are ``good enough'' \cite{dauphin2014identifying,choromanska2015loss}. In fact, it can be analytically shown that for some simple network architectures \cite{baldi1989neural,kawaguchi2016deep,lu2017depth}, there are no local minima and all critical points are either saddle points or the global minimum.


However, this theoretical picture, grounded on the topology of the loss function, has several shortcomings. First, the ultimate metric in machine learning is the generalisation accuracy rather than minimising the loss function, which only captures how well the model fits the training data. With vastly more parameters than samples, it is not difficult to completely fit the training data \cite{zhang2016understanding}. Therefore, theoretical results about the value of the loss function at local minima give limited insights about the effectiveness of SGD, which is its ability to find generalisable parameters. Second, it is known empirically that deeper minima have lower generalisation accuracy than shallower minima \cite{choromanska2015loss}. In fact, stopping SGD before complete convergence (``early stopping'') can yield parameters with higher generalisation accuracy compared to a converged SGD iteration \cite{prechelt2012early,duvenaud2016early}. Those empirical observations show that the local minimum that SGD finds is not only ``good enough'', but also better than deeper minima or even the global minimum that SGD misses.

Rather than focusing on the topology of the loss function, another strand of the literature focuses on the width of the minima. Pioneering works show that flatter minima lead to lower generalisation error \cite{hinton1993keeping,hochreiter1995simplifying,hochreiter1997flat}. The intuition is that a sharp minimum correspond to a more complex, likely overfitted, model because the parameters need to be accurately specified to define the minimum. Recent empirical studies have focused on the role of stochasticity on SGD: decreasing the batch size causes SGD to converge to wider minima and lead to better generalisation accuracy  \cite{keskar2016large, jastrzkebski2017three}, and in fact just adding Gaussian noise on top of SGD improves generalisation accuracy \cite{neelakantan2015adding}. An explicit term can also be added to bias SGD in favour of wider minima \cite{baldassi2016unreasonable,chaudhari2016entropy}. However, definitions of minima width in the literature are phenomenological and not reparametrization-invariant, leading to the critique that wide minima could be made arbitrarily sharp by simply rescaling the model parameters \cite{dinh2017sharp}.

In this paper, we will first analytically characterise the stochasticity in SGD. Our analysis will show how batch size affects stochasticity, and show that the noise has a non-trivial correlation structure which systematically biases SGD towards wide minima. We will then use a Bayesian approach to derive a correspondence between parameter inference and free energy minimisation in statistical physics with the degree of undersampling playing the role of temperature. Analogous to the energy-entropy competition in statistical physics, wide but shallow minima can be optimal if the system is undersampled. We will then apply our theoretical insight to study two prototypical non-convex models in machine learning: a numerical experiment of image classification using a deep neural network, and a linear neural network where we can derive analytical insights that explicitly relates out-of-sample error and entropy. 

\section{Stochastic gradient descent and entropy-energy competition}
\label{sgd}

To facilitate an analytic treatment of SGD, we can take the continuum limit and write Equation (\ref{SGD}) as a Langevin equation:
\begin{equation}
\frac{\mathrm{d}\boldsymbol{\theta}}{\mathrm{d}t} = - \nabla_{\boldsymbol{\theta} } L + \boldsymbol{\eta}(t),
\label{SGD_langevin_map}
\end{equation}
where
\begin{equation}
\boldsymbol{\eta}(t) = \nabla_{\boldsymbol{\theta} } \left[ \frac{1}{N} \sum_{i} l_i(\mathbf{x}_i,\boldsymbol{\theta})  -  \frac{1}{|B_t|} \sum_{i \in B_t} l_i(\mathbf{x}_i,\boldsymbol{\theta}) \right]
\end{equation}
is the stochastic noise. It is evident that $\left< \boldsymbol{\eta}(t)\right>=\mathbf{0}$, where the average is performed over batches of data. However, unlike thermal noise, the SGD noise has a non-trivial correlation structure. Extending the results of \cite{li2017batch}, for a constant batch size $|B_t| = b$ and $P\rightarrow \infty$, after some cumbersome algebra yields:
\begin{equation}
\small{
\left< \eta_\mu(t) \eta_\nu(t) \right> = \frac{1}{b} \left(1-\frac{b}{P} \right) \left[ \left< \frac{\partial l}{\partial \theta_\mu} \frac{\partial l}{\partial \theta_\nu}\right> - \left< \frac{\partial l}{\partial \theta_\mu} \right>\left< \frac{\partial l}{\partial \theta_\nu}\right> \right]},
\label{noise_structure}
\end{equation}
where $\left< f\right> = N^{-1} \sum_i f_i$ denotes the average over samples. Two insights can be gained from Equation (\ref{noise_structure}). First, the variance of the noise increases as $O(1/b)$. Therefore, decreasing the batch size increases the stochasticity, explaining the numerical observation of \cite{keskar2016large}. Second, the noise is highly anisotropic and biases SGD towards wide minima. To fix ideas, we consider the noise near a minima, where $\left< \partial l/ \partial \theta_\mu \right> \approx 0$ for all $\mu$. In the setting where the loss is the negative log probability of the parameters and data, the first term of Equation (\ref{noise_structure}) is simply the Fisher information, which equals the Hessian $H_{\mu \nu }= \left< \partial^2 l / (\partial \theta_\mu \partial \theta_\nu)\right>$. Suppose we reparametrise the system such that the Hessian is diagonal. Equation (\ref{noise_structure}) shows that ``stiff'' directions \cite{waterfall2006sloppy}, directions in the Hessian with larger curvature thus larger eigenvalue, is forced with a higher noise, whereas sloppier direction are forced with a lower noise by SGD. This is the opposite of many optimisation methods such as the Newton method where smaller steps are taken along stiffer directions. The net consequence of taking larger steps across stiff direction is that the algorithm can easily escape narrow basin and be trapped in wide basins.

Why might wide basins be generalisable? We consider a model regression problem, where
\begin{equation}
y_i = f(\mathbf{x}_i,\boldsymbol{\theta}^*)  + \epsilon_i,
\end{equation}
In general this $\epsilon$ noise can be thought of as arising from degrees of freedom which our model $f(\mathbf{x},\boldsymbol{\theta})$ does not capture because it does not have sufficient degrees of freedom to represent the true ``teacher''.  In other words, the noise reflects the approximation error of our model. $\boldsymbol{\theta}^*$ thus represent the best possible (unknown) parameters for the model to match the true teacher function. For simplicity and generality, we now approximate the distribution of the approximation error or noise to be Gaussian: $\epsilon_i \sim \mathcal{N}(0,\sigma_i)$, which is a reasonable assumption if it is composed by summing many random, independent components. We also assume the data points are independently sampled, so that the likelihood is:
\begin{equation}
p(\{\mathbf{x}_i \}_{i=1}^{P}| \boldsymbol{\theta})  \propto  \exp\left(- \sum_{i=1}^{P} \frac{\left( f(\mathbf{x}_i,\boldsymbol{\theta} \right)  - y_i)^2}{2 \sigma_i^2} \right).
\end{equation}
Applying a Gaussian prior $ p(\boldsymbol{\theta}) = \exp\left( -  \lambda ||\boldsymbol{\theta}||_2^2/2 \right)$ over parameters, the posterior is
\begin{equation}
p(\boldsymbol{\theta}| \{\mathbf{x}_i \}_{i=1}^{P}) = \frac{1}{Z} \exp\left( - \sum_{i=1}^{P} \frac{\left( f(\mathbf{x}_i,\boldsymbol{\theta} \right)  - y_i)^2}{2 \sigma_i^2} - \frac{ \lambda ||\boldsymbol{\theta}||_2^2}{2} \right).
\label{posterior}
\end{equation}
where $Z$ is a normalising constant. Statistical physics insights can be uncovered by first writing Equation (\ref{posterior}) in terms of \emph{intensive} quantities, i.e. quantitates that do not scale with the size of the data set $P$ or the number of parameters $N$. To this end, we define the mean error, which can be interpreted as an ``internal energy''
\begin{equation}
u(\boldsymbol{\theta}) = \frac{1}{P} \sum_{i=1}^{P} \frac{\left( f(\mathbf{x}_i,\boldsymbol{\theta} \right)  - y_i)^2}{2 \sigma_i^2} ,
\label{energy_term}
\end{equation}
and
\begin{equation}
h(\boldsymbol{\theta}) = \frac{\lambda}{N} \frac{||\boldsymbol{\theta}||_2^2}{2}.
\end{equation}
Given a new data point $\mathbf{\hat{x}}$, the goal is to predict the output $\hat{y}$. The expected $\hat{y}$ is given by:
\begin{equation}
\left< \hat{y} \right> = \int  p(\boldsymbol{\theta}| \{\mathbf{x}_i \}_{i=1}^{P}) f(\mathbf{\hat{x}},\boldsymbol{\theta}) \; \mathrm{d}{\boldsymbol{\theta}} .
\end{equation}
Machine learning problems where SGD have been successfully applied can be characterised by the asymptotic limit where number of data points $P \rightarrow \infty$, but the number of parameters is still vast and the data-to-parameter ratio $P/N = \alpha = O(1)$; this is often referred to as the high dimensional limit \cite{advani2013statistical}. For example, in image recognition challenges, the number of samples $P=O(10^{7})$ but the number of parameters in a deep neural network $N = O(10^{8})$ \cite{szegedy2016inception}. In this limit, it is justified to perform the Laplace approximation,
\begin{equation}
\left< \hat{y} \right> \approx \frac{1}{Z} \sum_q \frac{ \exp\left[- P (u(\boldsymbol{\theta}_q) - \frac{1}{\alpha} h(\boldsymbol{\theta}_q))\right]}{\sqrt{\det H(\boldsymbol{\theta}_q)}} f(\mathbf{\hat{x}},\boldsymbol{\theta}_q) .
\end{equation}
where $\boldsymbol{\theta}_q$ are the local minima of the loss function,
\begin{equation}
 \nabla_{\boldsymbol{\theta}} L(\{\mathbf{x}_i\}_{i=1}^{P},\boldsymbol{\theta}_q) = \nabla_{\boldsymbol{\theta}} \left( u(\boldsymbol{\theta_q}) - \frac{1}{\alpha}  h(\boldsymbol{\theta}_q) \right) = \mathbf{0},
\end{equation}
and $H(\boldsymbol{\theta}_q)$ is the Hessian matrix. The prefactor with the determinant of the Hessian can be written as
\begin{equation}
\frac{1}{\sqrt{\det H(\boldsymbol{\theta}_q)}} =  \exp\left( - \frac{1}{2} \sum_{i=1}^{N} \log \lambda_i(\boldsymbol{\theta}_q)  \right),
\end{equation}
where $\lambda_i$ are eigenvalues of the Hessian matrix. If the log eigenvalues are all of the same order of magnitude (note that the logarithm compresses scales), the sum is $O(N)$ and therefore the corresponding intensive quantity is
\begin{equation}
s(\boldsymbol{\theta}_q) = - \frac{1}{2 N} \sum_{i=1}^{N} \log \lambda_i(\boldsymbol{\theta}_q).
\label{entropy}
\end{equation}
The function $s(\boldsymbol{\theta}_q)$ can be interpreted as the entropy associated with that minimum. Geometrically, $s$ is related to the ``width'' of a minimum --- the wider a minimum is, the larger is the entropy function $s$ at that minimum. We note that we have assumed the eigenvalues to be non-zero --- this assumption is not necessarily true in deep neural networks that have more parameters than training data \cite{sagun2018empirical}, and we stress that a full analysis of basin volume \cite{martiniani2016structural} is required to characterise the entropy of flat minima. Moreover, we have made a further assumption that the numerical minimisation algorithms can actually find minima -- minimisation algorithms are often stuck in saddle points \cite{dauphin2014identifying}. Our numerical experiments below suggests that for singular Hessian or saddle points, our argument still holds qualitatively if we truncate the sum in Equation (\ref{entropy}) to include only large and positive eigenvalues.


All in all, $\left< \hat{y} \right>$ is approximately
\begin{equation}
\left< \hat{y} \right> \approx \frac{1}{Z} \sum_q e^{- P F(\boldsymbol{\theta}_q) } f(\mathbf{\hat{x}},\boldsymbol{\theta}_q),
\end{equation}
where
\begin{equation}
F(\boldsymbol{\theta}_q) = u(\boldsymbol{\theta}_q) + \frac{1}{\alpha} ( h(\boldsymbol{\theta}_q) -  s(\boldsymbol{\theta}_q))
\label{free_energy}
\end{equation}
is the effective free energy. When $P\rightarrow \infty$ and $\alpha$ is fixed, the minimum that dominates is \emph{not} the global minimum in $u(\boldsymbol{\theta})$, but rather the minimum with the lowest effective free energy. The free energy, Equation (\ref{free_energy}), has a transparent physical interpretation. The degree of undersampling $\frac{1}{\alpha}$ is analogous to the temperature, and the mean error and width of the minimum is analogous to the internal energy and entropy. The prior on the parameters, in the case of a Gaussian distribution, is analogous to an entropic spring that pulls the parameters towards the origin. Crucially, the definition of the entropy, which is derived from an asymptotic analysis of the posterior distribution, is based on the determinant of the Hessian  thus is invariant to reparameterization. This overcomes one critique in the literature on the correlation between basin width and model performance \cite{dinh2017sharp}.

We are now in a position to qualitatively understand why the Stochastic Gradient Descent algorithm works -- in the high dimensional limit, wide minima, in the sense of large $s$, are preferable over narrow ones even at the expense of higher error in-sample error $u$. The extent to which wide minima are preferable is dependent on the degree of undersampling and approximation noise. Wide local minima are of course easy to find via a gradient descent algorithm, and the noise in SGD specifically biases the search toward wider minima, thus partially explaining the unreasonable effectiveness of SGD.


\section{Numerical experiments}
Two testable predictions that emerge from our theory are: (1) The anisotropic noise in stochastic gradient descent forces it to converge to wider minima, and (2) subject to constant training error (``energy''), wider minima are more likely to have a low out-of-sample error compared to narrower minima. We will use a prototypical problem in machine learning, the CIFAR-10 challenge \cite{krizhevsky2009learning}, to test the extent to which those predictions are borne out in realistic datasets.

The CIFAR-10 dataset consists of 60000 $32\times32$ colour images of 10 different classes of objects (e.g. airplane, automobile, bird etc.), with 6000 images per class. The original machine learning problem is to parameterise a model that can classify an unseen image into one of those 10 classes. For numerical simplicity, we consider a stripped-down version of this problem where we only consider the binary classification problem of determining whether an image is an airplane or automobile. Moreover, we down sample the images to $10\times10$. Those simplifications are needed because numerically computing the Hessian for industrial scale models is numerically challenging.

We use a fully-connected neural network with three hidden layers and 10 units per layer. In mathematical terms, we flatten the image into a vector $\mathbf{x} \in \mathbb{R}^{100}$, and pose the model
\begin{equation}
y_i =  \sigma_1(\mathbf{W}_5 \sigma(\mathbf{W}_4 \sigma(\mathbf{W}_3 \sigma(\mathbf{W}_2\sigma_1(\mathbf{W}_1 \mathbf{x}_i )))))
\end{equation}
where $\mathbf{W}_1 \in \mathbb{R}^{10 \times 100}$, $\mathbf{W}_2 \in \mathbb{R}^{10 \times 10}$, $\mathbf{W}_3 \in \mathbb{R}^{10 \times 10}$, $\mathbf{W}_4 \in \mathbb{R}^{10 \times 10}$ and $\mathbf{W}_5 \in \mathbb{R}^{10 \times 2}$ are model parameters that we need to infer from the data, $\sigma(x) = \mathrm{max}(0,x)$ is known as the $\mathtt{relu}$ function in the literature and $\sigma_1(x) =(1+e^{-x})^{-1}$ is the sigmoid function . $y_i$ can be interpreted as a predicted probability -- the model is certain that the $i^{th}$ image is an airplane/automobile if $y_i=0/1$. The appropriate loss function for a classification problem is the so-call cross-entropy function
\begin{equation}
l_i = - t_i \log y_i - (1- t_i) \log(1 - y_i)
\end{equation}
where $t_i$ is the true label (0 for airplane, 1 for automobile) and $y_i$ is the predicted probability.

We have 12000 images of airplanes and automobiles in the dataset and 1320 parameters in our model. We will consider the case where we randomly choose 500 images as the training set. We restrict the number of images in the training data so that we can perfectly fit the model to training data (i.e. the ``energy'' $u=0$). The model suffers from overfitting, but the upshot is that we can focus solely on the entropy as the energy of every minimum is zero.

To verify whether the anisotropic noise in stochastic gradient descent forces the algorithm to converge to higher entropy minima, we compare the entropy of the critical points found via SGD with an alternative model of Langevin dynamics where the noise has the same magnitude as SGD but we remove the correlation structure. Concretely, we consider the Langevin dynamics
\begin{equation}
\boldsymbol{\theta}_{t+1} = \boldsymbol{\theta}_{t} - \eta G(\theta)   + \boldsymbol{\zeta}(t)
\label{SGD}
\end{equation}
where $G(\theta) = \nabla_{\boldsymbol{\theta} }\left(  L  + R(\boldsymbol{\theta}) \right)$ and the uncorrelated noise $\boldsymbol{\zeta}(t)$ is drawn from
\begin{equation}
\boldsymbol{\zeta}(t) \sim \mathcal{N}(0,\mathrm{diag}(\sigma_1(t), \sigma_2(t) \cdots, \sigma_p(t))),
\end{equation}
with
\begin{equation}
\sigma_{i}(t)^2 = \frac{1}{10} \sum_{m=1}^{10} (G_{m}^{(i)}(\boldsymbol{\theta})- \bar{G}^{(i)}(\boldsymbol{\theta}))^2, i=1,...,p
\end{equation}
and we estimate $\sigma_{i}^2$, the variance in the gradient in $i^{th}$ variable, by randomly splitting the sample into 10 minibatches of 50 samples each, estimate the minibatch gradient, and compute the variance of the gradient in the $i^{th}$ variable across the 10 minibatches.

To converge to a critical point, we run SGD with a minibatch size of 50 for 8000 epochs (full iteration through the dataset), followed by steepest descent minimisation until the gradient at each direction is less than $3 \times 10^{-5}$. Similarly, for the Langevin dynamics experiments, we run the modified Langevin dynamics for 80000 iterations, followed by steepest descent minimisation until the gradient at each direction is less than $3 \times 10^{-5}$.  For the full gradient descent without noise experiments, we run the full gradient descent for 80000 iterations and collect the data when the gradient at each direction is less than $3 \times 10^{-5}$. We start with a different initial condition for each run, with parameters drawn from a Gaus- sian distribution $\mathcal{N}(0,0.1)$. The learning rate $\eta$ for both SGD and our modified Langevin dynamics is 0.1. In both cases, a small regularisation of $\lambda =10^{-7}$ is used.

The Hessian matrix contains a small number of small negative eigenvalues (approximately 10\%), as well as positive eigenvalues close to zero. It is numerically challenging to differentiate whether those small negative or small positive eigenvalues are due to numerical noise masking a true minimum, a saddle point, or a degenerate minimum where the zero eigenvalues are masked by numerical noise. To make further progress, we use a phenomenological ``cleaning'' procedure where eigenvalues less than the magnitude of the most negative eigenvalue is ignored. This cleaning procedure is motivated by the intuition that the most negative eigenvalue gives an indication of the ``scale'' of numerical error.

\begin{figure}
\centering
\includegraphics[width=3.2in]{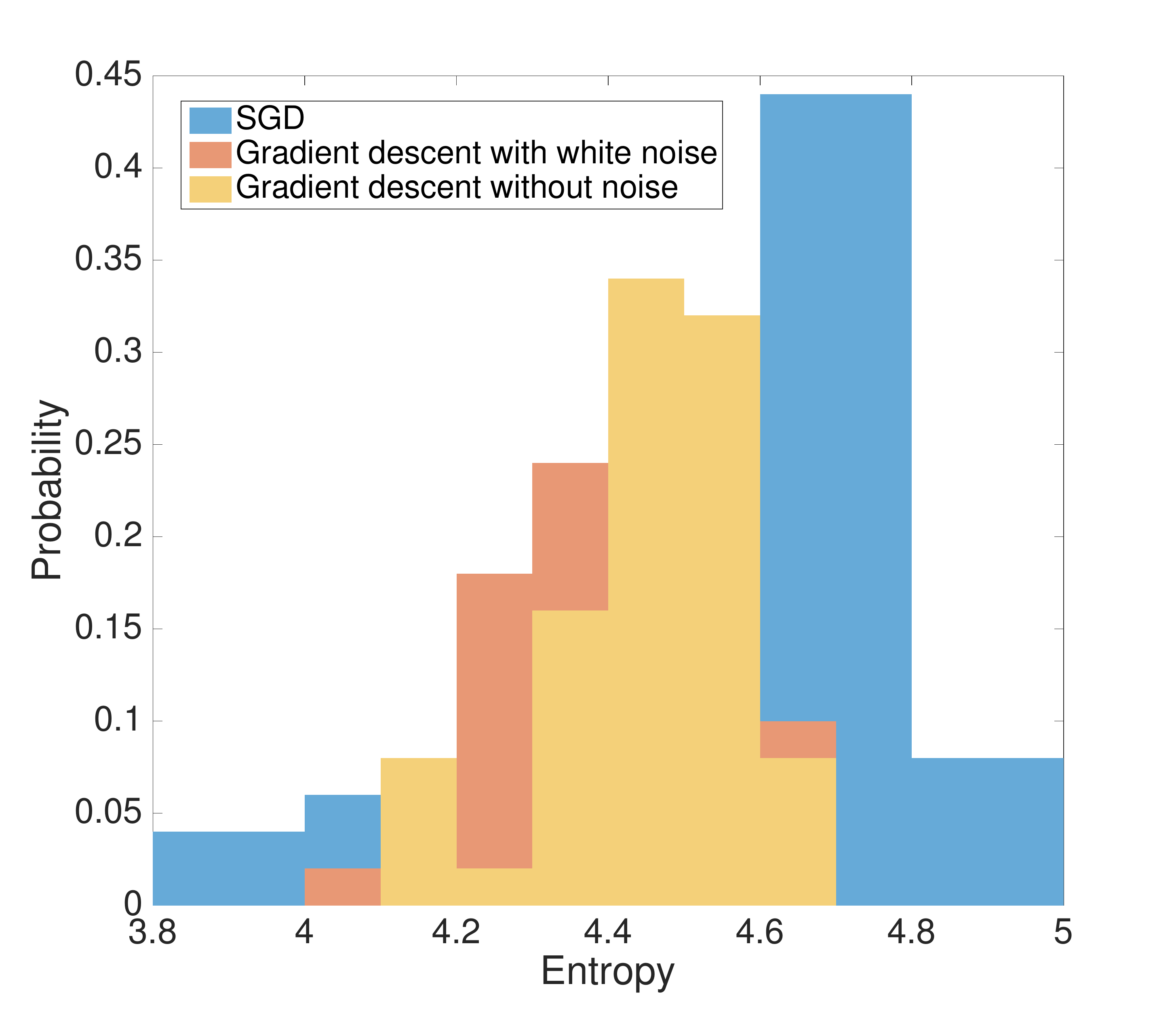}
\caption{SGD finds larger entropy solutions compared to Langevin dynamics or gradient descent. The plot shows a histogram of the entropy of the solutions found by the different optimization algorithms for a three-layer neural network trained on the binary classification problem of airplane and automobile images from CIFAR-10 datasets. The histogram is generated by 50 independent runs starting with random initial parameters.}
\label{histogram}
\end{figure}

Figure \ref{histogram} shows that the mean entropy of minima found via SGD is larger than the mean entropy found using Langevin dynamics with the same level of noise, confirming our hypothesis that the anisotropy of the noise biases SGD towards higher entropy minima.

\begin{figure}
\centering
\includegraphics[width=3in]{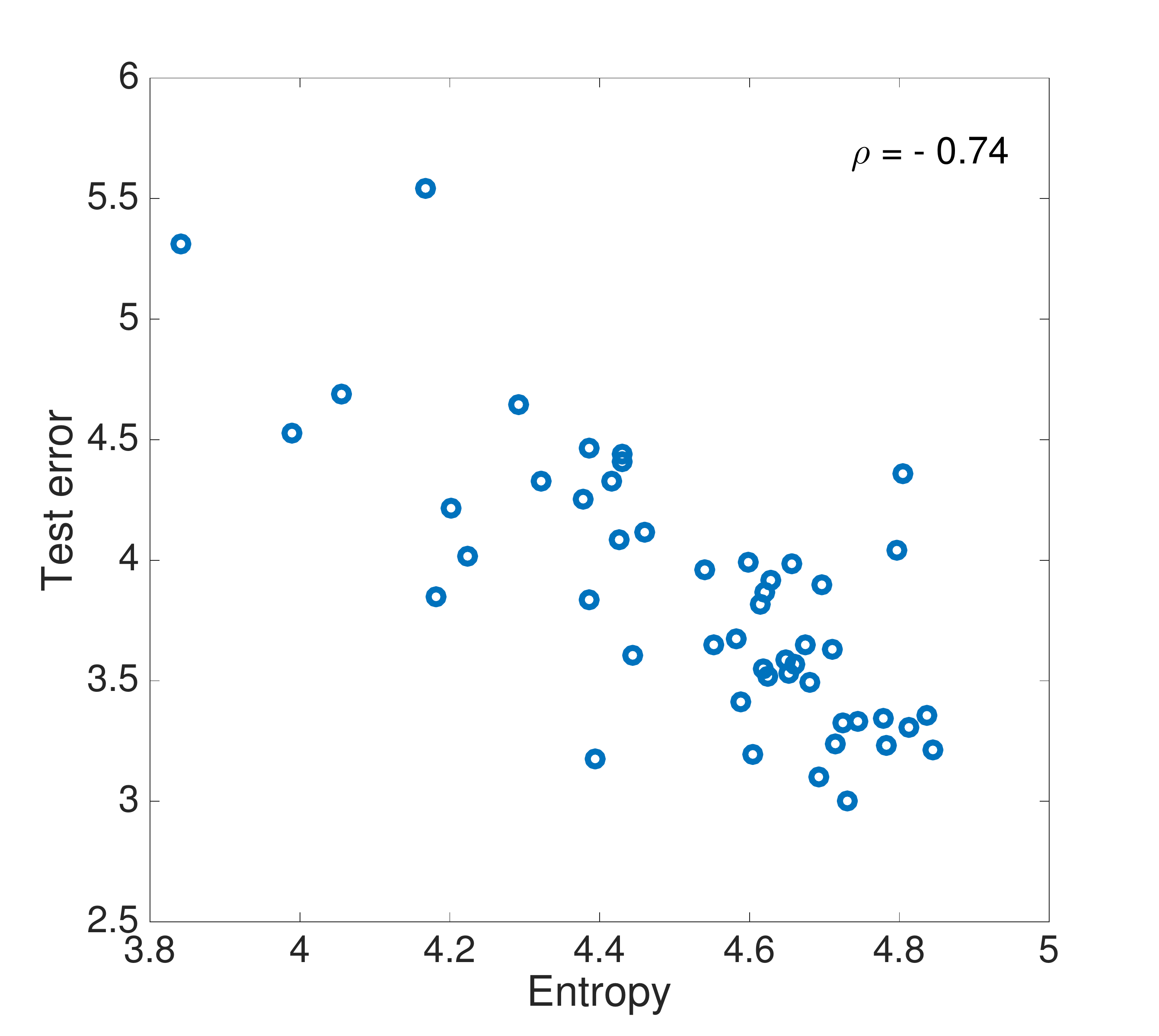}
\caption{Solutions with higher entropy are have lower out-of-sample error. We consider a a three-layer neural network trained and binary classification problem of airplane and automobile images discussed in the main text. The plot is generated by performing 50 independent runs of SGD starting from random initial weights. Note that the neural network is overparameterised, thus the in-sample error of the model is zero but the different solutions differs by their out-of-sample error.}
\label{entropy_gen}
\end{figure}

We next consider the correlation between entropy and the testing error, the error when the model is applied to images outside the training set. Using the same procedure discussed above, we locate critical points by running SGD for 8000 epochs followed by gradient descent minimisation. We compute the out-of-sample error by applying the model to 2000 images that are not in the training set. Figure \ref{entropy_gen} shows a discernible negative correlation between entropy and testing error. In other words, the larger the entropy is, the more likely the model can perform well outside the training set. Note that the error of the model in the test set is zero, therefore the only difference between the different solutions is entropy.

%

\section{Entropy in deep linear networks}

As another view onto the relationship between entropy and generalization performance, we examine the behavior of deep linear neural networks. Deep linear networks provide a simple model class that nevertheless retains important features of the learning problem faced in nonlinear deep networks. In particular, the optimization problem is nonconvex \cite{saxe2013exact}, and solution dynamics can be highly nonlinear \cite{saxe2013exact}. We consider a three-layer deep linear network which computes the output
\begin{equation}
\hat{y}_i =  \mathbf{W}_2\mathbf{W}_2 \mathbf{x}_i
\end{equation}
where $\mathbf{W}_1 \in \mathbb{R}^{N_h \times N_i}$ and $\mathbf{W}_2 \in \mathbb{R}^{1 \times N_h}$. We train these networks in a student-teacher setting (see e.g. \cite{Seung1992, advani_saxe_2017}), where a teacher labels random input examples $\mathbf{x}\sim\mathcal{N}(0,\frac{1}{N_i}\mathbf{I})$ as $y= \mathbf{\bar W}\mathbf{x}+\epsilon$. The teacher's parameters $\mathbf{\bar W}$ are drawn independently from a unit variance Gaussian distribution, and the noise variance (representing approximation error) is $\langle \epsilon^2\rangle=\sigma_e^2$. In this way a dataset of $P$ examples can be drawn and collected into an input matrix $\mathbf{X}\in \mathbb{R}^{N_i \times P}$ and target vector  $\mathbf{y}\in\mathbb{R}^{1 \times P}$. We assume a quadratic loss function so that the network aims to minimize $L = \sum_i(\hat{y}_i - y_i)^2$, and in our simulations, update weights with stochastic gradient descent.

\begin{figure}
\centering
\includegraphics[width=3.5in]{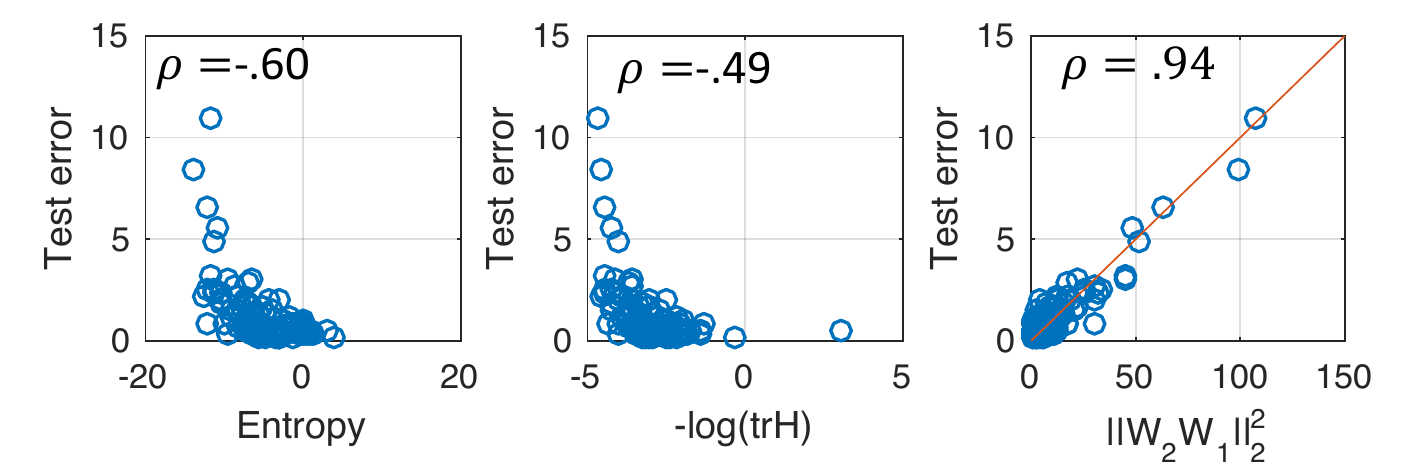}
\caption{Results of SGD in underdetermined deep linear networks, out-of-sample error correlates with entropy (left), the trace of the Hessian (middle) and the total weight norm (right). Correlation coefficients are given in each panel. We also demonstrate in the third panel the theoretical prediction \eqref{eq:egen_pred} for the average scaling of the generalization error with weight norm fits the data well. Other parameters: $N_i=10,N_h=7,P=5,\sigma_e=10^{-4}.$}
\label{linear_net}
\end{figure}

The generalization behavior of such networks is dependent on the amount of training data they receive \cite{advani_saxe_2017}. In the overdetermined case where data is plentiful ($P\geq N_i$), minimizers of the training error must compute identical input-output functions and attain identical generalization performance. However in the underdetermined case where $P<N_i$, minimizers of the training error all attain zero error on training samples, but need not compute identical input-output functions and therefore can generalize differently to new test samples. Intuitively, this is because the behavior of the input-output map is unconstrained in directions which contain no training data. We thus tested whether the correlation between entropy and generalization performance holds in the underdetermined regime. Fig.~\ref{linear_net} confirms this correlation for deep linear networks. Here as before, eigenvalues of the Hessian that are exactly zero have been discarded. We note that for this linear case, the number of nonzero eigenvalues is simply $\min(P,N_i)$ and hence by sorting the eigenvalues and taking the top $\min(P,N_i)$, issues of numerical precision are avoided.

 In the overdetermined setting, there is a manifold of global minima where the product of the weights is equal to the least squares solution, $ \mathbf{W}_2\mathbf{W}_1=\mathbf{\Sigma}^{yx}(\mathbf{\Sigma}^{x})^{-1}$ where  $\mathbf{\Sigma}^{yx}=\mathbf{y}\mathbf{X^T}$ and $\mathbf{\Sigma}^{x}=\mathbf{X}\mathbf{X}^T$ are the input-output covariance and input covariance respectively. All networks that attain this minimum must have the same input-output function and generalize identically. Even in this simple setting, however, the Hessian at every minimum has zero eigenvalues corresponding to directions in parameter space which point along the manifold of minima. In particular, a network with $N_h$ hidden units will have $N_h(N_i+1)-N_i$ zero eigenvalues.

 As the simplest example, a deep linear chain with scalar weights $w_2$ and $w_1$ (i.e. $N_i=N_h=1$) has Hessian
\begin{equation}
 H = \left[ \begin{array}{cc}
    w_2^2\sigma^x & 2w_2w_1\sigma^{x} - \sigma^{yx}  \\
2w_2w_1\sigma^{x} - \sigma^{yx} & w_1^2\sigma^{x}   \end{array} \right],
\end{equation}
yielding eigenvalues $\lambda_1 = w_2^2\sigma^x + w_1^2\sigma^{x}$ and $\lambda_2=0$ on the manifold of global minima. Hence the spectral norm of the Hessian coincides with the trace of the Hessian and is simply $\lambda_1$. Further, our entropy measure corresponds to taking the negative log of $\lambda_1$. The manifold of global minima is defined by the hyperbola $w_1w_2=\sigma^{yx}/\sigma^x$. Substituting this into $\lambda_1$, we have $\lambda_1=(\sigma^{yx})^2/(\sigma^xw_1^2)+w_1^2\sigma^{x}$ on the solution manifold, which retains a dependence on $w_1$. In particular, taking $w_1$ to infinity or zero will cause the entropy to go to infinity; but all points on the solution manifold implement the same function and generalize identically.  Thus a question remains: why do we observe a strong correlation between entropy and generalization error in typical training settings in spite of the parameter dependence of these Hessian-based metrics?

\begin{figure}
\centering
\includegraphics[width=3in]{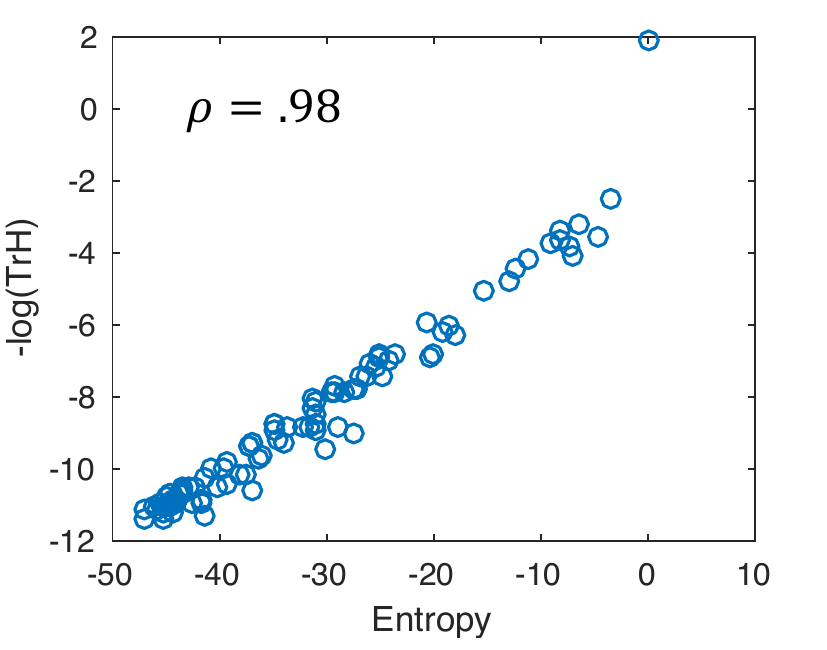}
\caption{The negative log of the trace of the Hessian correlates strongly with entropy.}
\label{entropy_trace}
\end{figure}

Some insight can be gained by examining the trace of the Hessian, which for these deep linear networks we find to be
\begin{equation}
	\textrm{tr} H = \left\|\mathbf{W}_2\right\|^2_2 \left\|\mathbf{X}\right\|_F^2 + \left\|\mathbf{W}_1\mathbf{X}\right\|_F^2,
	\label{trace_h}
\end{equation}
where $F$ denotes a Frobenius norm. As we show in Figure~\ref{entropy_trace}, the negative log of the trace of the Hessian closely tracks the entropy, and correlates about equally well with the generalization performance (Figure~\ref{linear_net} middle), so understanding the behavior of the trace can yield insight into the entropy measure as well. (We note, for instance, that in the linear chain considered above the entropy is exactly equal to $-\log(\textrm{tr} H)$). With this in mind, the reason for the correlation becomes evident: standard training procedures initialize the norm of each layer's weights to be approximately equal, and when so initialized, this persists through training. That is, under batch gradient descent learning (where every sample is used in each update $|B| = N$), the difference $\left\|\mathbf{W}_2 \right\|_2^2- \left\|\mathbf{W}_1 \right\|_F^2$ is an invariant of the dynamics. 

We can prove this fact by deriving the dynamics for this setting in the continuous limit by computing the derivatives of the loss function with respect to the weights. This yields the coupled differential equations:
\begin{eqnarray}
\tau \frac{d}{dt}\mathbf{W}_1 &= \mathbf{W}_2^T \left(\mathbf{\Sigma}^{yx} - \mathbf{W}_2\mathbf{W}_1\mathbf{\Sigma}^{xx}\right),\label{eq:dw1} \\
\tau \frac{d}{dt}\mathbf{W}_2 &=  \left(\mathbf{\Sigma}^{yx} - \mathbf{W}_2\mathbf{W}_1\mathbf{\Sigma}^{xx}\right) \mathbf{W}_1^T.\label{eq:dw0}
\end{eqnarray}
We can write our invariant as: 
\begin{equation}
\frac{d}{dt} \left(\left\|\mathbf{W}_2 \right\|_2^2- \left\|\mathbf{W}_1 \right\|_F^2 \right)= \frac{d}{dt} Tr\left(\mathbf{W}_2^T \mathbf{W}_2\right) - \frac{d}{dt}  Tr\left(\mathbf{W}_1 \mathbf{W}_1^T\right).
\end{equation}
Applying the derivative yields
\begin{equation}
Tr\left(\mathbf{W}_2^T \dot{\mathbf{W}}_2 - \dot{\mathbf{W}}_1 \mathbf{W}_1^T\right) + Tr\left(\mathbf{\dot{W}}_2^T \mathbf{W}_2 - \mathbf{W}_1 \dot{\mathbf{W}}_1^T\right)= 0,
\end{equation}
where the final equality holds by substituting \eqref{eq:dw1} and \eqref{eq:dw0} into the LHS of the expression above.

\begin{figure}
\centering
\includegraphics[width=3.5in]{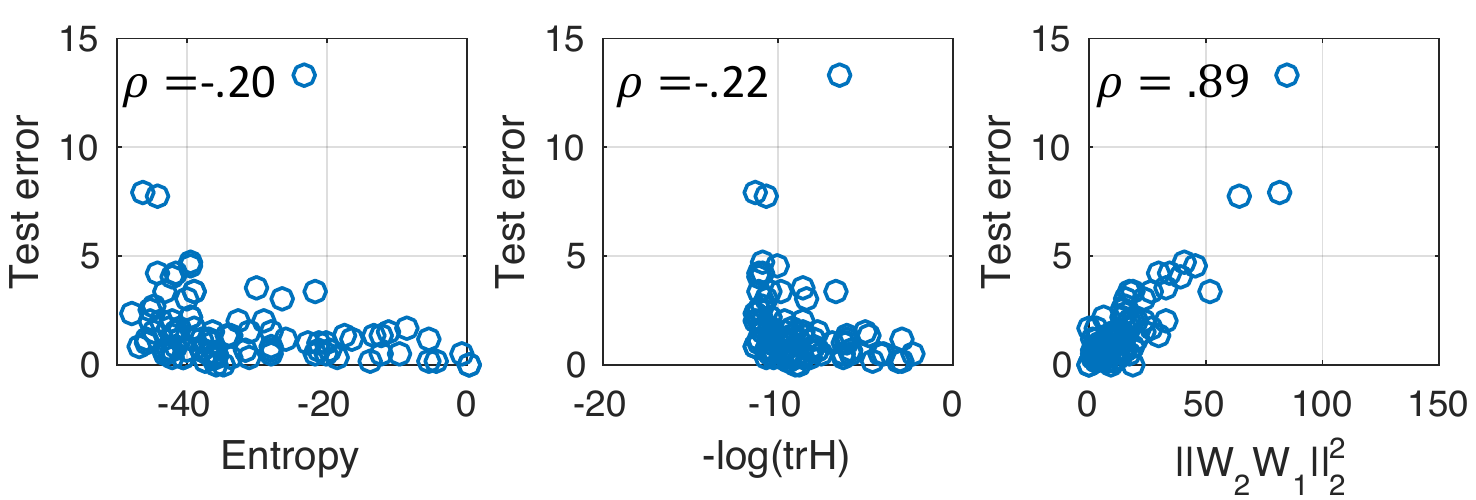}
\caption{Results of SGD in networks with extremely imbalanced initial weights ($\left\|\mathbf{W}_2 \right\|_2^2\geq \left\|\mathbf{W}_1 \right\|_F^2$ or vice versa), the correlation between out-of-sample error and entropy (left) or the trace of the Hessian (center) decreases, but the correlation with total weight norm remains unchanged (c.f. Fig.~\ref{linear_net}). Weights were initialized with random asymmetries between 1 and 100. Other parameters as in Fig.~\ref{linear_net}.}
\label{linear_net_assym}
\end{figure}

Batch gradient descent is equivalent to SGD for small learning rates and so this invariance can be expected to approximately hold in that setting. It follows that a balanced initial condition implicitly restricts the ultimate solution to be balanced as well, resolving the dependence of the trace and entropy metrics on the asymmetry in weight norms. Balanced solutions can still vary widely in norm, and from Equation (\ref{trace_h}), solutions with lower norm will have lower trace and entropy. For the linear networks considered here, the optimal out-of-sample error is attained by the minimum norm solution. Hence for deep linear networks, the correlation between entropy and generalization performance is stronger for balanced initial conditions (which are standard in practice). To demonstrate this, in Figure~\ref{linear_net_assym} we modify our initialization procedure to introduce asymmetry, yielding a reduced correlation.

The exact generalization error is given by $e_g=\frac{1}{N_i}\left\|\mathbf{\bar W}-\mathbf{W}_2\mathbf{W}_1\right\|_2^2+\sigma_e^2$. Now consider splitting these matrices into the component in the subspace of the training data, and the component perpendicular to the training data. We thus have 
\begin{equation}
e_g=\frac{1}{N_i}\left\|\mathbf{\bar W^\parallel}+\mathbf{\bar W^\perp}-\mathbf{W}_2\mathbf{W}_1^\parallel-\mathbf{W}_2\mathbf{W}_1^\perp\right\|_2^2+\sigma_e^2 .
\end{equation}
If training is successful, then $\mathbf{\bar W^\parallel}-\mathbf{W}_2\mathbf{W}_1^\parallel\approx 0$. Therefore 
\begin{equation}
e_g\approx \frac{1}{N_i}\left\|\mathbf{\bar W^\perp} -\mathbf{W}_2\mathbf{W}_1^\perp \right\|_2^2 + \sigma_e^2.
\end{equation}
Given $P$ examples, and $N_i$ input units, the dimension of the null space of the data covariance will be $N_i-P$. Also, there will be no correlation between $\bar{\mathbf{W}}^\perp$ and $\mathbf{W}_2 \mathbf{W}_1^{\perp}$ because the student weights are initialized randomly and are exposed to no information about the true weights in the direction orthogonal to the data. It follows that:

\begin{equation}
\left< e_g \right>  = \frac{1}{N_i}\left< \| \bar{\mathbf{W}}^\perp \|_2^2\right> + \frac{1}{N_i}\left< \| \mathbf{W}_2\mathbf{W}_1^\perp \|_2^2\right> + \sigma_e^2.
\end{equation}
To relate the previous expression to the norm of the product of the weights, we use the fact that
\begin{equation}
\|\mathbf{W}_2 \mathbf{W}_1\|_2^2 = \|\mathbf{W}_2 \mathbf{W}_1^\parallel + \mathbf{W}_2 \mathbf{W}_1^\perp\|_2^2 \approx  \| \bar{\mathbf{W}}^\parallel + \mathbf{W}_2 \mathbf{W}_1^\perp\|_2^2.
\end{equation}
Rearranging and averaging the expression above yields: 
\begin{equation}
\left< \|\mathbf{W}_2 \mathbf{W}_1^\perp\|_2^2\right>= -\left< \| \bar{\mathbf{W}}^\parallel \|_2^2\right> + \left<\|\mathbf{W}_2 \mathbf{W}_1\|_2^2\right> .
\end{equation}
Substituting this into our expression for generalization error yields:
\begin{equation}
\left< e_g \right>  = \frac{1}{N_i}\left(\left< \| \bar{\mathbf{W}}^\perp \|_2^2\right> - \left< \| \bar{\mathbf{W}}^\parallel \|_2^2\right>\right)+ \frac{1}{N_i}\left< \| \mathbf{W}_2\mathbf{W}_1 \|_2^2\right> + \sigma_e^2.
\label{eq:egen_pred}
\end{equation}
Thus, we have an expression for how the generalization performance of this student network is linearly related to the norm of its weights. The fit of this prediction may be seen in simulations in Figure \ref{linear_net}.

In the under-determined setting of a deep linear network, when the Hessian has many zero directions, we have shown that high entropy solutions are correlated with lower test error. In this simple setting, the reason for this connection is that the higher entropy solutions tend to have lower weight norms, and the weight norm is strongly correlated with the test error. In more complex non-linear networks, a similar mechanism may play a role, but there may also be additional causes behind why the entropy metric is related to generalization performance.

\section{Conclusion} 
We showed that the SGD algorithm, a workhorse minimisation algorithm in machine learning, preferentially finds minima that are wide because the noise is correlated and anisotropic. By performing a Laplace approximation, we show that the training error and log determinant of the Hessian matrix plays the role of energy and entropy in statistical physics, whilst the degree of undersampling plays the role of temperature. In the undersampled, ``high temperature'', regime the Bayes optimal parameters are determined by a balance between the training error and the basin width. This provides a physical justification of why stochastic gradient descent can train a model with low out-of-sample error despite almost certainly not finding the global minimum in the loss function. We tested the energy-entropy competition hypothesis with two prototypical machine learning models: a simple image classification problem and a linear neural network. The numerical experiments show that: (1) SGD locates minima with higher entropy compared to gradient descent with white noise (Langevin dynamics) or gradient descent without noise; (2) subject to constant training error, higher entropy solutions enjoy lower testing error. For the linear neural network model, we can analytically explain the relationship between entropy and test error. 

However, our analysis thus far focused on an harmonic expansion around minima, leading to a characterisation of entropy based on the log determinant of the Hessian matrix. There are two shortcomings with this characterisation. First, as is shown in the linear case, the Hessian matrix is singular for overparameterised systems, and zero eigenvalues are present also in overparametrised non-linear neural networks models \cite{sagun2018empirical}. Although we heuristically justified our phenomenological clipping of low eigenvalues in the regime of balanced weight initialisation, a more nuanced metric is needed to characterise flat minima with singular Hessian matrices. Second, the harmonic expansion approximation ignores higher order geometric features of the basins of attraction. The molecular physics community has developed a rich framework for describing energy landscapes \cite{wales2003energy}, and techniques from molecular simulations that can compute the basin volume of high dimensional energy landscape will likely yield significantly greater insights into the connection between basin volume and test error \cite{martiniani2016structural, ballard2017energy}. 

A perhaps more vexing issue is that neural networks seldom converge to a minima -- in fact, it can be shown analytically that the number of saddle points proliferates in high dimensional energy landscape functions \cite{dauphin2014identifying,choromanska2015loss}. Therefore, an open question is to go beyond characterising minima and develop a statistical framework that relate geometric properties of saddle points to the out-of-sample performance of machine learning models.

\begin{acknowledgements}
It is our pleasure to dedicate this work to Daan Frenkel on the occasion of his 70th birthday. Alpha Lee is particularly indebted to Daan for his continuing scientific inspiration and mentionship, from his first interactions with Daan as a teaching assistant in Park City to being colleagues in Cambridge. AAL is supported by the Winton Programme for the Physics of Sustainability, and AMS and MSA are supported by the Swartz Program in Theoretical Neuroscience.
\end{acknowledgements}

\bibliography{ref}

\begin{thebibliography}{38}%
\makeatletter
\providecommand \@ifxundefined [1]{%
 \@ifx{#1\undefined}
}%
\providecommand \@ifnum [1]{%
 \ifnum #1\expandafter \@firstoftwo
 \else \expandafter \@secondoftwo
 \fi
}%
\providecommand \@ifx [1]{%
 \ifx #1\expandafter \@firstoftwo
 \else \expandafter \@secondoftwo
 \fi
}%
\providecommand \natexlab [1]{#1}%
\providecommand \enquote  [1]{``#1''}%
\providecommand \bibnamefont  [1]{#1}%
\providecommand \bibfnamefont [1]{#1}%
\providecommand \citenamefont [1]{#1}%
\providecommand \href@noop [0]{\@secondoftwo}%
\providecommand \href [0]{\begingroup \@sanitize@url \@href}%
\providecommand \@href[1]{\@@startlink{#1}\@@href}%
\providecommand \@@href[1]{\endgroup#1\@@endlink}%
\providecommand \@sanitize@url [0]{\catcode `\\12\catcode `\$12\catcode
  `\&12\catcode `\#12\catcode `\^12\catcode `\_12\catcode `\%12\relax}%
\providecommand \@@startlink[1]{}%
\providecommand \@@endlink[0]{}%
\providecommand \url  [0]{\begingroup\@sanitize@url \@url }%
\providecommand \@url [1]{\endgroup\@href {#1}{\urlprefix }}%
\providecommand \urlprefix  [0]{URL }%
\providecommand \Eprint [0]{\href }%
\providecommand \doibase [0]{http://dx.doi.org/}%
\providecommand \selectlanguage [0]{\@gobble}%
\providecommand \bibinfo  [0]{\@secondoftwo}%
\providecommand \bibfield  [0]{\@secondoftwo}%
\providecommand \translation [1]{[#1]}%
\providecommand \BibitemOpen [0]{}%
\providecommand \bibitemStop [0]{}%
\providecommand \bibitemNoStop [0]{.\EOS\space}%
\providecommand \EOS [0]{\spacefactor3000\relax}%
\providecommand \BibitemShut  [1]{\csname bibitem#1\endcsname}%
\let\auto@bib@innerbib\@empty
\bibitem [{\citenamefont {Robbins}\ and\ \citenamefont
  {Monro}(1951)}]{robbins1951stochastic}%
  \BibitemOpen
  \bibfield  {author} {\bibinfo {author} {\bibfnamefont {H.}~\bibnamefont
  {Robbins}}\ and\ \bibinfo {author} {\bibfnamefont {S.}~\bibnamefont
  {Monro}},\ }\href@noop {} {\bibfield  {journal} {\bibinfo  {journal} {The
  annals of mathematical statistics}\ ,\ \bibinfo {pages} {400}} (\bibinfo
  {year} {1951})}\BibitemShut {NoStop}%
\bibitem [{\citenamefont {Bottou}(2010)}]{bottou2010large}%
  \BibitemOpen
  \bibfield  {author} {\bibinfo {author} {\bibfnamefont {L.}~\bibnamefont
  {Bottou}},\ }in\ \href@noop {} {\emph {\bibinfo {booktitle} {Proceedings of
  COMPSTAT'2010}}}\ (\bibinfo  {publisher} {Springer},\ \bibinfo {year}
  {2010})\ pp.\ \bibinfo {pages} {177--186}\BibitemShut {NoStop}%
\bibitem [{\citenamefont {Bottou}\ \emph {et~al.}(2016)\citenamefont {Bottou},
  \citenamefont {Curtis},\ and\ \citenamefont
  {Nocedal}}]{bottou2016optimization}%
  \BibitemOpen
  \bibfield  {author} {\bibinfo {author} {\bibfnamefont {L.}~\bibnamefont
  {Bottou}}, \bibinfo {author} {\bibfnamefont {F.~E.}\ \bibnamefont {Curtis}},
  \ and\ \bibinfo {author} {\bibfnamefont {J.}~\bibnamefont {Nocedal}},\
  }\href@noop {} {\bibfield  {journal} {\bibinfo  {journal} {arXiv preprint
  arXiv:1606.04838}\ } (\bibinfo {year} {2016})}\BibitemShut {NoStop}%
\bibitem [{\citenamefont {Krizhevsky}\ \emph {et~al.}(2012)\citenamefont
  {Krizhevsky}, \citenamefont {Sutskever},\ and\ \citenamefont
  {Hinton}}]{krizhevsky2012imagenet}%
  \BibitemOpen
  \bibfield  {author} {\bibinfo {author} {\bibfnamefont {A.}~\bibnamefont
  {Krizhevsky}}, \bibinfo {author} {\bibfnamefont {I.}~\bibnamefont
  {Sutskever}}, \ and\ \bibinfo {author} {\bibfnamefont {G.~E.}\ \bibnamefont
  {Hinton}},\ }in\ \href@noop {} {\emph {\bibinfo {booktitle} {Advances in
  neural information processing systems}}}\ (\bibinfo {year} {2012})\ pp.\
  \bibinfo {pages} {1097--1105}\BibitemShut {NoStop}%
\bibitem [{\citenamefont {Russakovsky}\ \emph {et~al.}(2015)\citenamefont
  {Russakovsky}, \citenamefont {Deng}, \citenamefont {Su}, \citenamefont
  {Krause}, \citenamefont {Satheesh}, \citenamefont {Ma}, \citenamefont
  {Huang}, \citenamefont {Karpathy}, \citenamefont {Khosla}, \citenamefont
  {Bernstein} \emph {et~al.}}]{russakovsky2015imagenet}%
  \BibitemOpen
  \bibfield  {author} {\bibinfo {author} {\bibfnamefont {O.}~\bibnamefont
  {Russakovsky}}, \bibinfo {author} {\bibfnamefont {J.}~\bibnamefont {Deng}},
  \bibinfo {author} {\bibfnamefont {H.}~\bibnamefont {Su}}, \bibinfo {author}
  {\bibfnamefont {J.}~\bibnamefont {Krause}}, \bibinfo {author} {\bibfnamefont
  {S.}~\bibnamefont {Satheesh}}, \bibinfo {author} {\bibfnamefont
  {S.}~\bibnamefont {Ma}}, \bibinfo {author} {\bibfnamefont {Z.}~\bibnamefont
  {Huang}}, \bibinfo {author} {\bibfnamefont {A.}~\bibnamefont {Karpathy}},
  \bibinfo {author} {\bibfnamefont {A.}~\bibnamefont {Khosla}}, \bibinfo
  {author} {\bibfnamefont {M.}~\bibnamefont {Bernstein}},  \emph {et~al.},\
  }\href@noop {} {\bibfield  {journal} {\bibinfo  {journal} {International
  Journal of Computer Vision}\ }\textbf {\bibinfo {volume} {115}},\ \bibinfo
  {pages} {211} (\bibinfo {year} {2015})}\BibitemShut {NoStop}%
\bibitem [{\citenamefont {LeCun}\ \emph {et~al.}(2015)\citenamefont {LeCun},
  \citenamefont {Bengio},\ and\ \citenamefont {Hinton}}]{lecun2015deep}%
  \BibitemOpen
  \bibfield  {author} {\bibinfo {author} {\bibfnamefont {Y.}~\bibnamefont
  {LeCun}}, \bibinfo {author} {\bibfnamefont {Y.}~\bibnamefont {Bengio}}, \
  and\ \bibinfo {author} {\bibfnamefont {G.}~\bibnamefont {Hinton}},\
  }\href@noop {} {\bibfield  {journal} {\bibinfo  {journal} {Nature}\ }\textbf
  {\bibinfo {volume} {521}},\ \bibinfo {pages} {436} (\bibinfo {year}
  {2015})}\BibitemShut {NoStop}%
\bibitem [{\citenamefont {Szegedy}\ \emph {et~al.}(2016)\citenamefont
  {Szegedy}, \citenamefont {Ioffe}, \citenamefont {Vanhoucke},\ and\
  \citenamefont {Alemi}}]{szegedy2016inception}%
  \BibitemOpen
  \bibfield  {author} {\bibinfo {author} {\bibfnamefont {C.}~\bibnamefont
  {Szegedy}}, \bibinfo {author} {\bibfnamefont {S.}~\bibnamefont {Ioffe}},
  \bibinfo {author} {\bibfnamefont {V.}~\bibnamefont {Vanhoucke}}, \ and\
  \bibinfo {author} {\bibfnamefont {A.}~\bibnamefont {Alemi}},\ }\href@noop {}
  {\bibfield  {journal} {\bibinfo  {journal} {arXiv preprint arXiv:1602.07261}\
  } (\bibinfo {year} {2016})}\BibitemShut {NoStop}%
\bibitem [{\citenamefont {Silver}\ \emph {et~al.}(2016)\citenamefont {Silver},
  \citenamefont {Huang}, \citenamefont {Maddison}, \citenamefont {Guez},
  \citenamefont {Sifre}, \citenamefont {Van Den~Driessche}, \citenamefont
  {Schrittwieser}, \citenamefont {Antonoglou}, \citenamefont {Panneershelvam},
  \citenamefont {Lanctot} \emph {et~al.}}]{silver2016mastering}%
  \BibitemOpen
  \bibfield  {author} {\bibinfo {author} {\bibfnamefont {D.}~\bibnamefont
  {Silver}}, \bibinfo {author} {\bibfnamefont {A.}~\bibnamefont {Huang}},
  \bibinfo {author} {\bibfnamefont {C.~J.}\ \bibnamefont {Maddison}}, \bibinfo
  {author} {\bibfnamefont {A.}~\bibnamefont {Guez}}, \bibinfo {author}
  {\bibfnamefont {L.}~\bibnamefont {Sifre}}, \bibinfo {author} {\bibfnamefont
  {G.}~\bibnamefont {Van Den~Driessche}}, \bibinfo {author} {\bibfnamefont
  {J.}~\bibnamefont {Schrittwieser}}, \bibinfo {author} {\bibfnamefont
  {I.}~\bibnamefont {Antonoglou}}, \bibinfo {author} {\bibfnamefont
  {V.}~\bibnamefont {Panneershelvam}}, \bibinfo {author} {\bibfnamefont
  {M.}~\bibnamefont {Lanctot}},  \emph {et~al.},\ }\href@noop {} {\bibfield
  {journal} {\bibinfo  {journal} {Nature}\ }\textbf {\bibinfo {volume} {529}},\
  \bibinfo {pages} {484} (\bibinfo {year} {2016})}\BibitemShut {NoStop}%
\bibitem [{\citenamefont {Fyodorov}\ and\ \citenamefont
  {Williams}(2007)}]{fyodorov2007replica}%
  \BibitemOpen
  \bibfield  {author} {\bibinfo {author} {\bibfnamefont {Y.~V.}\ \bibnamefont
  {Fyodorov}}\ and\ \bibinfo {author} {\bibfnamefont {I.}~\bibnamefont
  {Williams}},\ }\href@noop {} {\bibfield  {journal} {\bibinfo  {journal}
  {Journal of Statistical Physics}\ }\textbf {\bibinfo {volume} {129}},\
  \bibinfo {pages} {1081} (\bibinfo {year} {2007})}\BibitemShut {NoStop}%
\bibitem [{\citenamefont {Bray}\ and\ \citenamefont
  {Dean}(2007)}]{bray2007statistics}%
  \BibitemOpen
  \bibfield  {author} {\bibinfo {author} {\bibfnamefont {A.~J.}\ \bibnamefont
  {Bray}}\ and\ \bibinfo {author} {\bibfnamefont {D.~S.}\ \bibnamefont
  {Dean}},\ }\href@noop {} {\bibfield  {journal} {\bibinfo  {journal} {Physical
  review letters}\ }\textbf {\bibinfo {volume} {98}},\ \bibinfo {pages}
  {150201} (\bibinfo {year} {2007})}\BibitemShut {NoStop}%
\bibitem [{\citenamefont {Dauphin}\ \emph {et~al.}(2014)\citenamefont
  {Dauphin}, \citenamefont {Pascanu}, \citenamefont {Gulcehre}, \citenamefont
  {Cho}, \citenamefont {Ganguli},\ and\ \citenamefont
  {Bengio}}]{dauphin2014identifying}%
  \BibitemOpen
  \bibfield  {author} {\bibinfo {author} {\bibfnamefont {Y.~N.}\ \bibnamefont
  {Dauphin}}, \bibinfo {author} {\bibfnamefont {R.}~\bibnamefont {Pascanu}},
  \bibinfo {author} {\bibfnamefont {C.}~\bibnamefont {Gulcehre}}, \bibinfo
  {author} {\bibfnamefont {K.}~\bibnamefont {Cho}}, \bibinfo {author}
  {\bibfnamefont {S.}~\bibnamefont {Ganguli}}, \ and\ \bibinfo {author}
  {\bibfnamefont {Y.}~\bibnamefont {Bengio}},\ }in\ \href@noop {} {\emph
  {\bibinfo {booktitle} {Advances in neural information processing systems}}}\
  (\bibinfo {year} {2014})\ pp.\ \bibinfo {pages} {2933--2941}\BibitemShut
  {NoStop}%
\bibitem [{\citenamefont {Choromanska}\ \emph {et~al.}(2015)\citenamefont
  {Choromanska}, \citenamefont {Henaff}, \citenamefont {Mathieu}, \citenamefont
  {Arous},\ and\ \citenamefont {LeCun}}]{choromanska2015loss}%
  \BibitemOpen
  \bibfield  {author} {\bibinfo {author} {\bibfnamefont {A.}~\bibnamefont
  {Choromanska}}, \bibinfo {author} {\bibfnamefont {M.}~\bibnamefont {Henaff}},
  \bibinfo {author} {\bibfnamefont {M.}~\bibnamefont {Mathieu}}, \bibinfo
  {author} {\bibfnamefont {G.~B.}\ \bibnamefont {Arous}}, \ and\ \bibinfo
  {author} {\bibfnamefont {Y.}~\bibnamefont {LeCun}},\ }in\ \href@noop {}
  {\emph {\bibinfo {booktitle} {AISTATS}}}\ (\bibinfo {year}
  {2015})\BibitemShut {NoStop}%
\bibitem [{\citenamefont {Baldi}\ and\ \citenamefont
  {Hornik}(1989)}]{baldi1989neural}%
  \BibitemOpen
  \bibfield  {author} {\bibinfo {author} {\bibfnamefont {P.}~\bibnamefont
  {Baldi}}\ and\ \bibinfo {author} {\bibfnamefont {K.}~\bibnamefont {Hornik}},\
  }\href@noop {} {\bibfield  {journal} {\bibinfo  {journal} {Neural networks}\
  }\textbf {\bibinfo {volume} {2}},\ \bibinfo {pages} {53} (\bibinfo {year}
  {1989})}\BibitemShut {NoStop}%
\bibitem [{\citenamefont {Kawaguchi}(2016)}]{kawaguchi2016deep}%
  \BibitemOpen
  \bibfield  {author} {\bibinfo {author} {\bibfnamefont {K.}~\bibnamefont
  {Kawaguchi}},\ }in\ \href@noop {} {\emph {\bibinfo {booktitle} {Advances in
  Neural Information Processing Systems}}}\ (\bibinfo {year} {2016})\ pp.\
  \bibinfo {pages} {586--594}\BibitemShut {NoStop}%
\bibitem [{\citenamefont {Lu}\ and\ \citenamefont
  {Kawaguchi}(2017)}]{lu2017depth}%
  \BibitemOpen
  \bibfield  {author} {\bibinfo {author} {\bibfnamefont {H.}~\bibnamefont
  {Lu}}\ and\ \bibinfo {author} {\bibfnamefont {K.}~\bibnamefont {Kawaguchi}},\
  }\href@noop {} {\bibfield  {journal} {\bibinfo  {journal} {arXiv preprint
  arXiv:1702.08580}\ } (\bibinfo {year} {2017})}\BibitemShut {NoStop}%
\bibitem [{\citenamefont {Zhang}\ \emph {et~al.}(2016)\citenamefont {Zhang},
  \citenamefont {Bengio}, \citenamefont {Hardt}, \citenamefont {Recht},\ and\
  \citenamefont {Vinyals}}]{zhang2016understanding}%
  \BibitemOpen
  \bibfield  {author} {\bibinfo {author} {\bibfnamefont {C.}~\bibnamefont
  {Zhang}}, \bibinfo {author} {\bibfnamefont {S.}~\bibnamefont {Bengio}},
  \bibinfo {author} {\bibfnamefont {M.}~\bibnamefont {Hardt}}, \bibinfo
  {author} {\bibfnamefont {B.}~\bibnamefont {Recht}}, \ and\ \bibinfo {author}
  {\bibfnamefont {O.}~\bibnamefont {Vinyals}},\ }\href@noop {} {\bibfield
  {journal} {\bibinfo  {journal} {arXiv preprint arXiv:1611.03530}\ } (\bibinfo
  {year} {2016})}\BibitemShut {NoStop}%
\bibitem [{\citenamefont {Prechelt}(2012)}]{prechelt2012early}%
  \BibitemOpen
  \bibfield  {author} {\bibinfo {author} {\bibfnamefont {L.}~\bibnamefont
  {Prechelt}},\ }in\ \href@noop {} {\emph {\bibinfo {booktitle} {Neural
  Networks: Tricks of the Trade}}}\ (\bibinfo  {publisher} {Springer},\
  \bibinfo {year} {2012})\ pp.\ \bibinfo {pages} {53--67}\BibitemShut {NoStop}%
\bibitem [{\citenamefont {Duvenaud}\ \emph {et~al.}(2016)\citenamefont
  {Duvenaud}, \citenamefont {Maclaurin},\ and\ \citenamefont
  {Adams}}]{duvenaud2016early}%
  \BibitemOpen
  \bibfield  {author} {\bibinfo {author} {\bibfnamefont {D.}~\bibnamefont
  {Duvenaud}}, \bibinfo {author} {\bibfnamefont {D.}~\bibnamefont {Maclaurin}},
  \ and\ \bibinfo {author} {\bibfnamefont {R.}~\bibnamefont {Adams}},\ }in\
  \href@noop {} {\emph {\bibinfo {booktitle} {Artificial Intelligence and
  Statistics}}}\ (\bibinfo {year} {2016})\ pp.\ \bibinfo {pages}
  {1070--1077}\BibitemShut {NoStop}%
\bibitem [{\citenamefont {Hinton}\ and\ \citenamefont
  {Van~Camp}(1993)}]{hinton1993keeping}%
  \BibitemOpen
  \bibfield  {author} {\bibinfo {author} {\bibfnamefont {G.~E.}\ \bibnamefont
  {Hinton}}\ and\ \bibinfo {author} {\bibfnamefont {D.}~\bibnamefont
  {Van~Camp}},\ }in\ \href@noop {} {\emph {\bibinfo {booktitle} {Proceedings of
  the sixth annual conference on Computational learning theory}}}\ (\bibinfo
  {organization} {ACM},\ \bibinfo {year} {1993})\ pp.\ \bibinfo {pages}
  {5--13}\BibitemShut {NoStop}%
\bibitem [{\citenamefont {Hochreiter}\ and\ \citenamefont
  {Schmidhuber}(1995)}]{hochreiter1995simplifying}%
  \BibitemOpen
  \bibfield  {author} {\bibinfo {author} {\bibfnamefont {S.}~\bibnamefont
  {Hochreiter}}\ and\ \bibinfo {author} {\bibfnamefont {J.}~\bibnamefont
  {Schmidhuber}},\ }\href@noop {} {\bibfield  {journal} {\bibinfo  {journal}
  {Advances in Neural Information Processing Systems}\ ,\ \bibinfo {pages}
  {529}} (\bibinfo {year} {1995})}\BibitemShut {NoStop}%
\bibitem [{\citenamefont {Hochreiter}\ and\ \citenamefont
  {Schmidhuber}(1997)}]{hochreiter1997flat}%
  \BibitemOpen
  \bibfield  {author} {\bibinfo {author} {\bibfnamefont {S.}~\bibnamefont
  {Hochreiter}}\ and\ \bibinfo {author} {\bibfnamefont {J.}~\bibnamefont
  {Schmidhuber}},\ }\href@noop {} {\bibfield  {journal} {\bibinfo  {journal}
  {Neural Computation}\ }\textbf {\bibinfo {volume} {9}},\ \bibinfo {pages} {1}
  (\bibinfo {year} {1997})}\BibitemShut {NoStop}%
\bibitem [{\citenamefont {Keskar}\ \emph {et~al.}(2016)\citenamefont {Keskar},
  \citenamefont {Mudigere}, \citenamefont {Nocedal}, \citenamefont
  {Smelyanskiy},\ and\ \citenamefont {Tang}}]{keskar2016large}%
  \BibitemOpen
  \bibfield  {author} {\bibinfo {author} {\bibfnamefont {N.~S.}\ \bibnamefont
  {Keskar}}, \bibinfo {author} {\bibfnamefont {D.}~\bibnamefont {Mudigere}},
  \bibinfo {author} {\bibfnamefont {J.}~\bibnamefont {Nocedal}}, \bibinfo
  {author} {\bibfnamefont {M.}~\bibnamefont {Smelyanskiy}}, \ and\ \bibinfo
  {author} {\bibfnamefont {P.~T.~P.}\ \bibnamefont {Tang}},\ }\href@noop {}
  {\bibfield  {journal} {\bibinfo  {journal} {arXiv preprint arXiv:1609.04836}\
  } (\bibinfo {year} {2016})}\BibitemShut {NoStop}%
\bibitem [{\citenamefont {Jastrzebski}\ \emph {et~al.}(2017)\citenamefont
  {Jastrzebski}, \citenamefont {Kenton}, \citenamefont {Arpit}, \citenamefont
  {Ballas}, \citenamefont {Fischer}, \citenamefont {Bengio},\ and\
  \citenamefont {Storkey}}]{jastrzkebski2017three}%
  \BibitemOpen
  \bibfield  {author} {\bibinfo {author} {\bibfnamefont {S.}~\bibnamefont
  {Jastrzebski}}, \bibinfo {author} {\bibfnamefont {Z.}~\bibnamefont {Kenton}},
  \bibinfo {author} {\bibfnamefont {D.}~\bibnamefont {Arpit}}, \bibinfo
  {author} {\bibfnamefont {N.}~\bibnamefont {Ballas}}, \bibinfo {author}
  {\bibfnamefont {A.}~\bibnamefont {Fischer}}, \bibinfo {author} {\bibfnamefont
  {Y.}~\bibnamefont {Bengio}}, \ and\ \bibinfo {author} {\bibfnamefont
  {A.}~\bibnamefont {Storkey}},\ }\href@noop {} {\bibfield  {journal} {\bibinfo
   {journal} {arXiv preprint arXiv:1711.04623}\ } (\bibinfo {year}
  {2017})}\BibitemShut {NoStop}%
\bibitem [{\citenamefont {Neelakantan}\ \emph {et~al.}(2015)\citenamefont
  {Neelakantan}, \citenamefont {Vilnis}, \citenamefont {Le}, \citenamefont
  {Sutskever}, \citenamefont {Kaiser}, \citenamefont {Kurach},\ and\
  \citenamefont {Martens}}]{neelakantan2015adding}%
  \BibitemOpen
  \bibfield  {author} {\bibinfo {author} {\bibfnamefont {A.}~\bibnamefont
  {Neelakantan}}, \bibinfo {author} {\bibfnamefont {L.}~\bibnamefont {Vilnis}},
  \bibinfo {author} {\bibfnamefont {Q.~V.}\ \bibnamefont {Le}}, \bibinfo
  {author} {\bibfnamefont {I.}~\bibnamefont {Sutskever}}, \bibinfo {author}
  {\bibfnamefont {L.}~\bibnamefont {Kaiser}}, \bibinfo {author} {\bibfnamefont
  {K.}~\bibnamefont {Kurach}}, \ and\ \bibinfo {author} {\bibfnamefont
  {J.}~\bibnamefont {Martens}},\ }\href@noop {} {\bibfield  {journal} {\bibinfo
   {journal} {arXiv preprint arXiv:1511.06807}\ } (\bibinfo {year}
  {2015})}\BibitemShut {NoStop}%
\bibitem [{\citenamefont {Baldassi}\ \emph {et~al.}(2016)\citenamefont
  {Baldassi}, \citenamefont {Borgs}, \citenamefont {Chayes}, \citenamefont
  {Ingrosso}, \citenamefont {Lucibello}, \citenamefont {Saglietti},\ and\
  \citenamefont {Zecchina}}]{baldassi2016unreasonable}%
  \BibitemOpen
  \bibfield  {author} {\bibinfo {author} {\bibfnamefont {C.}~\bibnamefont
  {Baldassi}}, \bibinfo {author} {\bibfnamefont {C.}~\bibnamefont {Borgs}},
  \bibinfo {author} {\bibfnamefont {J.~T.}\ \bibnamefont {Chayes}}, \bibinfo
  {author} {\bibfnamefont {A.}~\bibnamefont {Ingrosso}}, \bibinfo {author}
  {\bibfnamefont {C.}~\bibnamefont {Lucibello}}, \bibinfo {author}
  {\bibfnamefont {L.}~\bibnamefont {Saglietti}}, \ and\ \bibinfo {author}
  {\bibfnamefont {R.}~\bibnamefont {Zecchina}},\ }\href@noop {} {\bibfield
  {journal} {\bibinfo  {journal} {Proceedings of the National Academy of
  Sciences}\ }\textbf {\bibinfo {volume} {113}},\ \bibinfo {pages} {E7655}
  (\bibinfo {year} {2016})}\BibitemShut {NoStop}%
\bibitem [{\citenamefont {Chaudhari}\ \emph {et~al.}(2016)\citenamefont
  {Chaudhari}, \citenamefont {Choromanska}, \citenamefont {Soatto},\ and\
  \citenamefont {LeCun}}]{chaudhari2016entropy}%
  \BibitemOpen
  \bibfield  {author} {\bibinfo {author} {\bibfnamefont {P.}~\bibnamefont
  {Chaudhari}}, \bibinfo {author} {\bibfnamefont {A.}~\bibnamefont
  {Choromanska}}, \bibinfo {author} {\bibfnamefont {S.}~\bibnamefont {Soatto}},
  \ and\ \bibinfo {author} {\bibfnamefont {Y.}~\bibnamefont {LeCun}},\
  }\href@noop {} {\bibfield  {journal} {\bibinfo  {journal} {arXiv preprint
  arXiv:1611.01838}\ } (\bibinfo {year} {2016})}\BibitemShut {NoStop}%
\bibitem [{\citenamefont {Dinh}\ \emph {et~al.}(2017)\citenamefont {Dinh},
  \citenamefont {Pascanu}, \citenamefont {Bengio},\ and\ \citenamefont
  {Bengio}}]{dinh2017sharp}%
  \BibitemOpen
  \bibfield  {author} {\bibinfo {author} {\bibfnamefont {L.}~\bibnamefont
  {Dinh}}, \bibinfo {author} {\bibfnamefont {R.}~\bibnamefont {Pascanu}},
  \bibinfo {author} {\bibfnamefont {S.}~\bibnamefont {Bengio}}, \ and\ \bibinfo
  {author} {\bibfnamefont {Y.}~\bibnamefont {Bengio}},\ }\href@noop {}
  {\bibfield  {journal} {\bibinfo  {journal} {arXiv preprint arXiv:1703.04933}\
  } (\bibinfo {year} {2017})}\BibitemShut {NoStop}%
\bibitem [{\citenamefont {Li}\ \emph {et~al.}(2017)\citenamefont {Li},
  \citenamefont {Li}, \citenamefont {Qian},\ and\ \citenamefont
  {Liu}}]{li2017batch}%
  \BibitemOpen
  \bibfield  {author} {\bibinfo {author} {\bibfnamefont {C.~J.}\ \bibnamefont
  {Li}}, \bibinfo {author} {\bibfnamefont {L.}~\bibnamefont {Li}}, \bibinfo
  {author} {\bibfnamefont {J.}~\bibnamefont {Qian}}, \ and\ \bibinfo {author}
  {\bibfnamefont {J.-G.}\ \bibnamefont {Liu}},\ }\href@noop {} {\bibfield
  {journal} {\bibinfo  {journal} {arXiv preprint arXiv:1705.07562}\ } (\bibinfo
  {year} {2017})}\BibitemShut {NoStop}%
\bibitem [{\citenamefont {Waterfall}\ \emph {et~al.}(2006)\citenamefont
  {Waterfall}, \citenamefont {Casey}, \citenamefont {Gutenkunst}, \citenamefont
  {Brown}, \citenamefont {Myers}, \citenamefont {Brouwer}, \citenamefont
  {Elser},\ and\ \citenamefont {Sethna}}]{waterfall2006sloppy}%
  \BibitemOpen
  \bibfield  {author} {\bibinfo {author} {\bibfnamefont {J.~J.}\ \bibnamefont
  {Waterfall}}, \bibinfo {author} {\bibfnamefont {F.~P.}\ \bibnamefont
  {Casey}}, \bibinfo {author} {\bibfnamefont {R.~N.}\ \bibnamefont
  {Gutenkunst}}, \bibinfo {author} {\bibfnamefont {K.~S.}\ \bibnamefont
  {Brown}}, \bibinfo {author} {\bibfnamefont {C.~R.}\ \bibnamefont {Myers}},
  \bibinfo {author} {\bibfnamefont {P.~W.}\ \bibnamefont {Brouwer}}, \bibinfo
  {author} {\bibfnamefont {V.}~\bibnamefont {Elser}}, \ and\ \bibinfo {author}
  {\bibfnamefont {J.~P.}\ \bibnamefont {Sethna}},\ }\href@noop {} {\bibfield
  {journal} {\bibinfo  {journal} {Physical review letters}\ }\textbf {\bibinfo
  {volume} {97}},\ \bibinfo {pages} {150601} (\bibinfo {year}
  {2006})}\BibitemShut {NoStop}%
\bibitem [{\citenamefont {Advani}\ \emph {et~al.}(2013)\citenamefont {Advani},
  \citenamefont {Lahiri},\ and\ \citenamefont
  {Ganguli}}]{advani2013statistical}%
  \BibitemOpen
  \bibfield  {author} {\bibinfo {author} {\bibfnamefont {M.}~\bibnamefont
  {Advani}}, \bibinfo {author} {\bibfnamefont {S.}~\bibnamefont {Lahiri}}, \
  and\ \bibinfo {author} {\bibfnamefont {S.}~\bibnamefont {Ganguli}},\
  }\href@noop {} {\bibfield  {journal} {\bibinfo  {journal} {Journal of
  Statistical Mechanics: Theory and Experiment}\ }\textbf {\bibinfo {volume}
  {2013}},\ \bibinfo {pages} {P03014} (\bibinfo {year} {2013})}\BibitemShut
  {NoStop}%
\bibitem [{\citenamefont {Sagun}\ \emph {et~al.}(2018)\citenamefont {Sagun},
  \citenamefont {Evci}, \citenamefont {Guney}, \citenamefont {Dauphin},\ and\
  \citenamefont {Bottou}}]{sagun2018empirical}%
  \BibitemOpen
  \bibfield  {author} {\bibinfo {author} {\bibfnamefont {L.}~\bibnamefont
  {Sagun}}, \bibinfo {author} {\bibfnamefont {U.}~\bibnamefont {Evci}},
  \bibinfo {author} {\bibfnamefont {V.~U.}\ \bibnamefont {Guney}}, \bibinfo
  {author} {\bibfnamefont {Y.}~\bibnamefont {Dauphin}}, \ and\ \bibinfo
  {author} {\bibfnamefont {L.}~\bibnamefont {Bottou}},\ }\href@noop {}
  {\bibfield  {journal} {\bibinfo  {journal} {arXiv:1706.04454}\ } (\bibinfo
  {year} {2018})}\BibitemShut {NoStop}%
\bibitem [{\citenamefont {Martiniani}\ \emph {et~al.}(2016)\citenamefont
  {Martiniani}, \citenamefont {Schrenk}, \citenamefont {Stevenson},
  \citenamefont {Wales},\ and\ \citenamefont
  {Frenkel}}]{martiniani2016structural}%
  \BibitemOpen
  \bibfield  {author} {\bibinfo {author} {\bibfnamefont {S.}~\bibnamefont
  {Martiniani}}, \bibinfo {author} {\bibfnamefont {K.~J.}\ \bibnamefont
  {Schrenk}}, \bibinfo {author} {\bibfnamefont {J.~D.}\ \bibnamefont
  {Stevenson}}, \bibinfo {author} {\bibfnamefont {D.~J.}\ \bibnamefont
  {Wales}}, \ and\ \bibinfo {author} {\bibfnamefont {D.}~\bibnamefont
  {Frenkel}},\ }\href@noop {} {\bibfield  {journal} {\bibinfo  {journal}
  {Physical Review E}\ }\textbf {\bibinfo {volume} {94}},\ \bibinfo {pages}
  {031301} (\bibinfo {year} {2016})}\BibitemShut {NoStop}%
\bibitem [{\citenamefont {Krizhevsky}\ and\ \citenamefont
  {Hinton}(2009)}]{krizhevsky2009learning}%
  \BibitemOpen
  \bibfield  {author} {\bibinfo {author} {\bibfnamefont {A.}~\bibnamefont
  {Krizhevsky}}\ and\ \bibinfo {author} {\bibfnamefont {G.}~\bibnamefont
  {Hinton}},\ }\href@noop {} {\emph {\bibinfo {title} {Learning multiple layers
  of features from tiny images}}},\ \bibinfo {type} {Tech. Rep.}\ (\bibinfo
  {year} {2009})\BibitemShut {NoStop}%
\bibitem [{\citenamefont {Saxe}\ \emph {et~al.}(2013)\citenamefont {Saxe},
  \citenamefont {McClelland},\ and\ \citenamefont {Ganguli}}]{saxe2013exact}%
  \BibitemOpen
  \bibfield  {author} {\bibinfo {author} {\bibfnamefont {A.~M.}\ \bibnamefont
  {Saxe}}, \bibinfo {author} {\bibfnamefont {J.~L.}\ \bibnamefont
  {McClelland}}, \ and\ \bibinfo {author} {\bibfnamefont {S.}~\bibnamefont
  {Ganguli}},\ }\href@noop {} {\bibfield  {journal} {\bibinfo  {journal} {arXiv
  preprint arXiv:1312.6120}\ } (\bibinfo {year} {2013})}\BibitemShut {NoStop}%
\bibitem [{\citenamefont {Seung}\ \emph {et~al.}(1992)\citenamefont {Seung},
  \citenamefont {Sompolinsky},\ and\ \citenamefont {Tishby}}]{Seung1992}%
  \BibitemOpen
  \bibfield  {author} {\bibinfo {author} {\bibfnamefont {H.}~\bibnamefont
  {Seung}}, \bibinfo {author} {\bibfnamefont {H.}~\bibnamefont {Sompolinsky}},
  \ and\ \bibinfo {author} {\bibfnamefont {N.}~\bibnamefont {Tishby}},\ }\href
  {\doibase 10.1103/PhysRevA.45.6056} {\bibfield  {journal} {\bibinfo
  {journal} {Physical Review A}\ }\textbf {\bibinfo {volume} {45}},\ \bibinfo
  {pages} {6056} (\bibinfo {year} {1992})}\BibitemShut {NoStop}%
\bibitem [{\citenamefont {Advani}\ and\ \citenamefont
  {Saxe}(2017)}]{advani_saxe_2017}%
  \BibitemOpen
  \bibfield  {author} {\bibinfo {author} {\bibfnamefont {M.~S.}\ \bibnamefont
  {Advani}}\ and\ \bibinfo {author} {\bibfnamefont {A.~M.}\ \bibnamefont
  {Saxe}},\ }\href@noop {} {\bibfield  {journal} {\bibinfo  {journal} {arXiv
  preprint arXiv:1710.03667}\ } (\bibinfo {year} {2017})}\BibitemShut {NoStop}%
\bibitem [{\citenamefont {Wales}(2003)}]{wales2003energy}%
  \BibitemOpen
  \bibfield  {author} {\bibinfo {author} {\bibfnamefont {D.}~\bibnamefont
  {Wales}},\ }\href@noop {} {\emph {\bibinfo {title} {Energy landscapes:
  Applications to clusters, biomolecules and glasses}}}\ (\bibinfo  {publisher}
  {Cambridge University Press},\ \bibinfo {year} {2003})\BibitemShut {NoStop}%
\bibitem [{\citenamefont {Ballard}\ \emph {et~al.}(2017)\citenamefont
  {Ballard}, \citenamefont {Das}, \citenamefont {Martiniani}, \citenamefont
  {Mehta}, \citenamefont {Sagun}, \citenamefont {Stevenson},\ and\
  \citenamefont {Wales}}]{ballard2017energy}%
  \BibitemOpen
  \bibfield  {author} {\bibinfo {author} {\bibfnamefont {A.~J.}\ \bibnamefont
  {Ballard}}, \bibinfo {author} {\bibfnamefont {R.}~\bibnamefont {Das}},
  \bibinfo {author} {\bibfnamefont {S.}~\bibnamefont {Martiniani}}, \bibinfo
  {author} {\bibfnamefont {D.}~\bibnamefont {Mehta}}, \bibinfo {author}
  {\bibfnamefont {L.}~\bibnamefont {Sagun}}, \bibinfo {author} {\bibfnamefont
  {J.~D.}\ \bibnamefont {Stevenson}}, \ and\ \bibinfo {author} {\bibfnamefont
  {D.~J.}\ \bibnamefont {Wales}},\ }\href@noop {} {\bibfield  {journal}
  {\bibinfo  {journal} {Physical Chemistry Chemical Physics}\ }\textbf
  {\bibinfo {volume} {19}},\ \bibinfo {pages} {12585} (\bibinfo {year}
  {2017})}\BibitemShut {NoStop}%
\end{thebibliography}%

\end{document}